\documentclass[sigconf]{acmart}
\setcopyright{none}
\renewcommand\footnotetextcopyrightpermission[1]{}
\AtBeginDocument{%
  }

\setcopyright{acmlicensed}
\copyrightyear{2018}
\acmYear{2018}
\acmDOI{XXXXXXX.XXXXXXX}
\acmConference[Conference acronym 'XX]{Make sure to enter the correct
  conference title from your rights confirmation email}{June 03--05,
  2018}{Woodstock, NY}
\acmISBN{978-1-4503-XXXX-X/2018/06}

\usepackage{booktabs}
\usepackage{multirow}
\usepackage[table]{xcolor}
\usepackage{array}
\usepackage{dsfont}
\usepackage{arydshln}
\usepackage{subfig}
\definecolor{myblue}{RGB}{143,170,220}
\definecolor{mygreen}{RGB}{169,209,142}
\definecolor{myorange}{RGB}{244,177,131}
\definecolor{mypurple}{RGB}{188,170,218}
\definecolor{1}{RGB}{222,169,0}
\definecolor{2}{RGB}{255,215,87}
\definecolor{3}{RGB}{255,231,155}

\begin{document}

\title{Are VLMs Lost Between Sky and Space? LinkS$^2$Bench for UAV-Satellite Dynamic Cross-View Spatial Intelligence}

\author{Dian Liu}
\email{dianliuxd@stu.xidian.edu.cn}
\affiliation{%
  \institution{Xidian University}
  \country{China}
}

\author{Jie Feng}
\authornote{Corresponding author.}
\email{jiefeng0109@163.com}
\affiliation{%
  \institution{Xidian University}
  \country{China}
}

\author{Di Li}
\email{dili@stu.xidian.edu.cn}
\affiliation{%
  \institution{Xidian University}
  \country{China}
}

\author{Yuhui Zheng}
\email{zhengyh@vip.126.com}
\affiliation{%
  \institution{Qinghai Normal University}
  \country{China}
}

\author{Guanbin Li}
\email{liguanbin@mail.sysu.edu.cn}
\affiliation{%
  \institution{Sun Yat-sen University}
  \country{China}
}

\author{Weisheng Dong}
\email{wsdong@mail.xidian.edu.cn}
\affiliation{%
  \institution{Xidian University}
  \country{China}
}

\author{Guangming Shi}
\email{gmshi@xidian.edu.cn}
\affiliation{%
  \institution{Xidian University}
  \country{China}
}

\renewcommand{\shortauthors}{Trovato et al.}

\begin{abstract}
Synergistic spatial intelligence between UAVs and satellites is indispensable for emergency response and security operations, as it uniquely integrates macro-scale global coverage with dynamic, real-time local perception. However, the capacity of Vision-Language Models (VLMs) to master this complex interplay remains largely unexplored. 
This gap persists primarily because existing benchmarks are confined to isolated Unmanned Aerial Vehicle (UAV) videos or static satellite imagery, failing to evaluate the dynamic local-to-global spatial mapping essential for comprehensive cross-view reasoning. 
To bridge this gap, we introduce LinkS$^2$Bench, the first comprehensive benchmark designed to evaluate VLMs' wide-area, dynamic cross-view spatial intelligence. LinkS$^2$Bench links 1,022 minutes of dynamic UAV footage with high-resolution satellite imagery covering over 200 km². Through an LMM-assisted pipeline and rigorous human annotation, we constructed 17.9k high-quality question-answer pairs comprising 12 fine-grained tasks across four dimensions: perception, localization, relation, and reasoning.
Evaluations of 18 representative VLMs reveal a substantial gap compared to human baselines, identifying accurate cross-view dynamic alignment as the critical bottleneck. To alleviate this, we design a Cross-View Alignment Adapter, demonstrating that explicit alignment significantly improves model performance. Furthermore, fine-tuning experiments underscore the potential of LinkS$^2$Bench in advancing VLM adaptation for complex spatial reasoning.

\end{abstract}





\begin{teaserfigure}
    \centering
    \subfloat[Benchmark Motivation and Overview]{\includegraphics[width=0.6\linewidth]{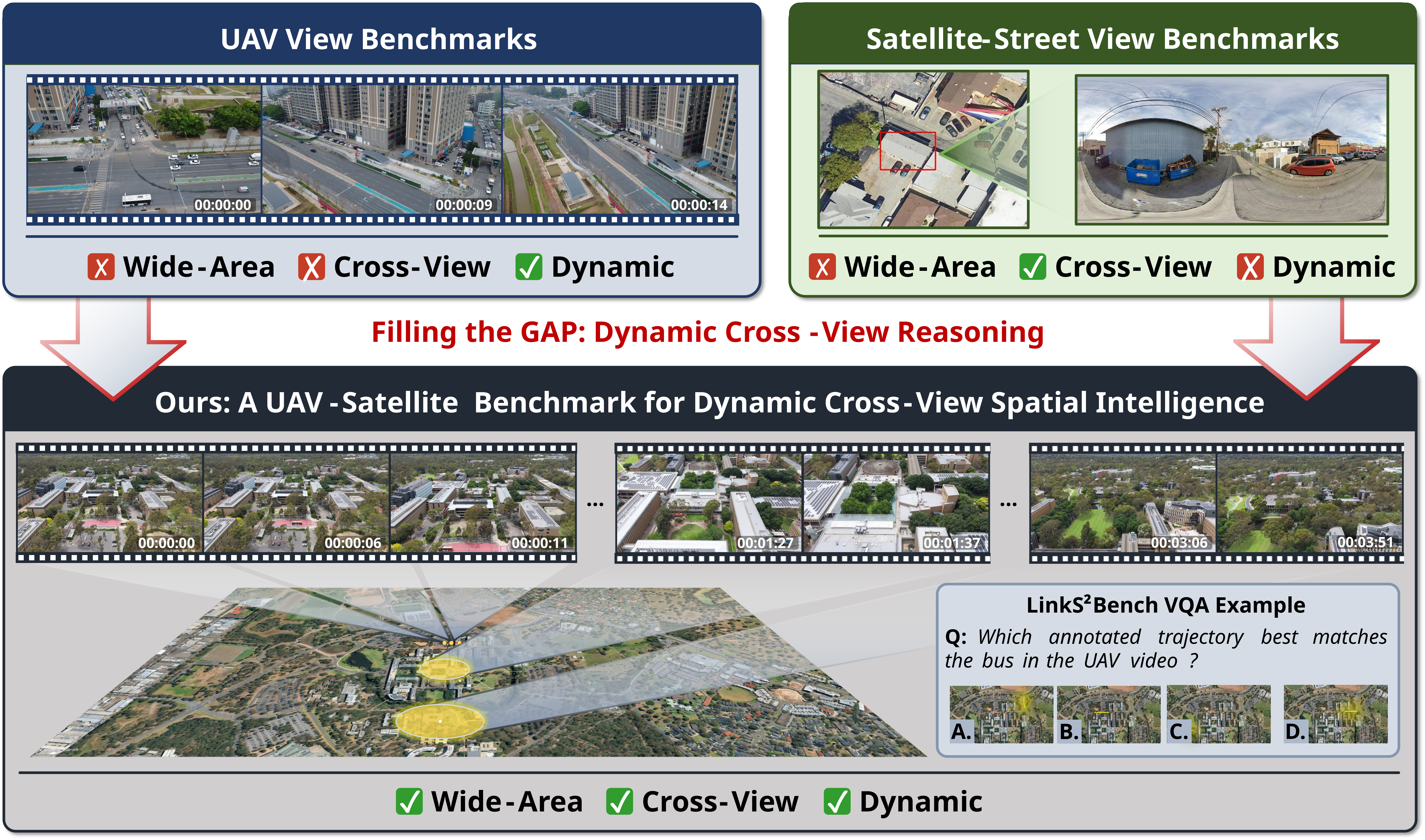}
    \label{fig:motivation}}
    \hfill
    \subfloat[Model Performance Comparison]{\includegraphics[width=0.35\linewidth]{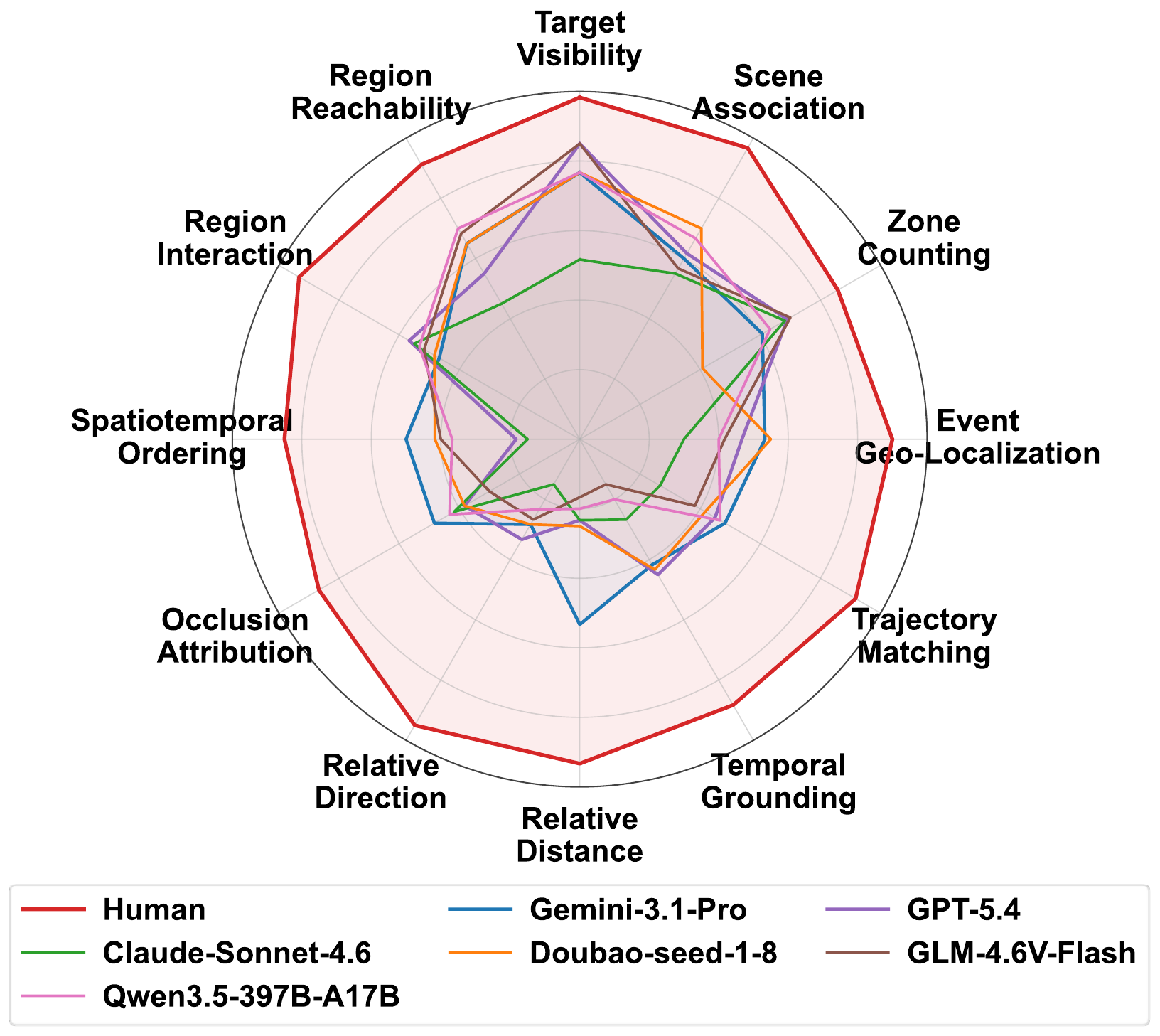}
    \label{fig:performance_comparison}}
    \caption{Overview of LinkS$^2$Bench. (a) Compared with existing benchmarks, LinkS$^2$Bench uniquely combines wide-area coverage, UAV-satellite cross-view reasoning, and dynamic spatiotemporal understanding. (b) Evaluation of representative VLMs against human performance across 12 fine-grained tasks reveals a substantial gap in cross-view spatial intelligence.}
\end{teaserfigure}


\maketitle

\section{Introduction}

In mission-critical applications such as autonomous navigation, dynamic target monitoring, and wide-area search-and-rescue, the synergy between Unmanned Aerial Vehicles (UAVs) and satellite imagery is indispensable ~\cite{wang2024towards, zhao2025cityeqa, ferrag2025uavbench, wang2025uavscenes}. Within this framework, UAVs deliver high-resolution, dynamic local observations, whereas satellites furnish a stable, wide-area global context. However, achieving genuine spatial intelligence in such dual-view environments is exceptionally challenging ~\cite{yin2025spatial, wang2025site, zhu2025cospace, daxberger2025mm}, as it requires vision models to seamlessly map transient local details onto a static global reference. This process necessitates overcoming drastic scale discrepancies and viewpoint transformations to achieve precise dynamic cross-view spatial alignment ~\cite{xu2024uav, ju2025video2bev, ji2025mmgeo}.

While recent Vision-Language Models (VLMs) have achieved remarkable success in general 2D image recognition and semantic visual question answering, their capacity for sophisticated cross-view spatial reasoning remains severely unexplored~\cite{tang2025roboafford, song2025robospatial, chen2024spatialvlm, zhou2025they, lee2025perspective, cheng2024spatialrgpt, gholami2025spatial}. This deficiency is largely attributable to the absence of a comprehensive benchmark that captures the complexities of UAV-satellite collaborative scenarios. General spatial visual question answering (VQA) benchmarks are largely confined to indoor or synthetic environments~\cite{li2025viewspatial, zhang2025video, zhao2025cityeqa, xu2026citycube, yang2025thinking, yang2025mmsi, dang2025rynnec, li2025sti, lee2025perspective}, failing to reflect the macro-scale physical constraints and extreme pose transformations inherent in real-world applications. Furthermore, as illustrated in Figure ~\ref{fig:motivation}, traditional UAV-view benchmarks~\cite{zhao2025urbanvideo, lin2025openvln, lee2025citynav, gao2025openfly, xiao2025uav, wang2025uav} capture rich dynamic sequences but remain constrained to isolated viewpoints, lacking the cross-view reference necessary for global map alignment. In contrast, existing satellite–street-view benchmarks~\cite{zhou2025urbench, feng2025urbanllava, ye2025cross, ye2025leveraging} successfully establish cross-view correspondences yet are inherently static, devoid of the dynamic local perception required for real-time monitoring. Consequently, there remains a dearth of benchmark capable of evaluating a VLM's ability to bridge dynamic local observations with a wide-area global context.

To address these limitations, we introduce \textbf{LinkS$^2$Bench}, the first benchmark specifically engineered to evaluate VLMs' cross-view spatial intelligence in dynamic UAV-satellite settings. Built from 1,022 minutes of real-world UAV footage and high-resolution satellite imagery covering over 200 km\textsuperscript{2}, LinkS\textsuperscript{2}Bench comprises 17.9k high-quality VQA pairs curated through an LMM-assisted pipeline and human annotation. Compared with existing benchmarks, 
LinkS\textsuperscript{2}Bench systematically evaluates spatial intelligence across three critical dimensions:
\begin{itemize}
\item \textbf{Wide-area spatial localization} tests a model's ability to go beyond simple object detection and achieve precise regional or coordinate grounding within expansive, georeferenced environments;
\item \textbf{Dynamic cross-view perception} examines the model's capacity to align transient, motion-rich UAV observations with stable satellite imagery, bridging domain gaps and perspective transformations;
\item \textbf{Spatiotemporal reasoning} evaluates the capacity to continuously monitor entity states and infer evolving interactions and causal dynamics across sequential video frames within a unified global spatial context.
\end{itemize}

Building upon the LinkS$^2$Bench benchmark, we conduct extensive evaluations of representative VLMs, which reveal a substantial performance gap compared to human baselines (Figure ~\ref{fig:performance_comparison}). Through a granular failure analysis, we identify \textbf{cross-view dynamic misalignment} as the primary bottleneck for current models. To address this, we propose a \textbf{Cross-View Alignment Adapter (CVAA)}, demonstrating that explicitly enhancing spatial alignment significantly boosts performance. Furthermore, supervised fine-tuning on LinkS$^2$Bench yields substantial gains, validating its utility as both a rigorous diagnostic benchmark and a valuable supervision source for model adaptation.

Our main contributions are summarized as follows:
\begin{enumerate}
\item We propose LinkS$^2$Bench, a comprehensive VQA benchmark for dynamic UAV-satellite cross-view scenarios, designed to for evaluate VLMs' cross-view spatial intelligence in terms of wide-area spatial localization, dynamic cross-view perception, and spatiotemporal reasoning.
\item We build LinkS$^2$Bench on a real-world UAV-satellite data foundation comprising 1,500 UAV videos and 43,273 satellite images from 16 cities worldwide, with over 800 hours of manual effort devoted to data collection and processing.
\item We develop a systematic task framework for dynamic cross-view spatial intelligence, organizing LinkS$^2$Bench into 12 fine-grained tasks across four capability dimensions: perception, localization, relation, and reasoning, and construct 17.9k high-quality VQA pairs through an LMM-assisted semi-automatic pipeline and over 1,000 hours of human annotation.
\item Through extensive experiments on 18 representative VLMs, we identify cross-view dynamic misalignment as the core bottleneck in current models. To address this, we propose the CVAA that significantly improves spatial alignment, and further validate the LinkS\textsuperscript{2}Bench's utility for model adaptation through supervised fine-tuning.
\end{enumerate}

\begin{table*}[t]
  \caption{Comparison of LinkS$^2$Bench with representative spatial intelligence benchmarks.}
  \label{tab:after-2-2}
  \centering
  \footnotesize
  \setlength{\tabcolsep}{10pt}
  \renewcommand{\arraystretch}{1.1}
  \begin{tabular}{lcccccc}
    \toprule
    Benchmark & View Setting & Input Modality & Dynamic & Cross-view & \# Tasks & \# QA Pairs \\
    \midrule
    VSI-Bench (CVPR 2025) ~\cite{yang2025thinking} & Ground & Video & $\checkmark$ & $\times$ & 8 & 5.0k \\
    All-Angles-Bench (AAAI 2026) ~\cite{yeh2025seeing} & Ground & Image & $\times$ & $\times$ & 6 & 2.1k \\
    Open3D-VQA (ACM MM 2025) ~\cite{zhang2025open3d} & UAV & Image & $\times$ & $\times$ & 7 & 73.3k \\
    AirCopBench (AAAI 2026) ~\cite{zha2025aircopbench} & UAV & Image & $\times$ & $\times$ & 14 & 14.6k \\
    SpatialSky-Bench (CVPR 2026) ~\cite{zhang2025your} & UAV & Image & $\times$ & $\times$ & 13 & 1.0k \\
    UrbanVideo-Bench (ACL 2025) ~\cite{zhao2025urbanvideo} & UAV & Video & $\checkmark$ & $\times$ & 16 & 5.2k \\
    UrBench (AAAI 2025) ~\cite{zhou2025urbench} & Street-Satellite & Image & $\times$ & $\checkmark$ & 14 & 11.6k \\
    \midrule
    Ours & UAV-Satellite & Video \& Image & $\checkmark$ & $\checkmark$ & 12 & 17.9k \\
    \bottomrule
  \end{tabular}
  \Description{Comparison of benchmarks by view setting, input modality, cross-view ability, dynamic scenes, and number of QA pairs.}
  \label{tab:comparison}
\end{table*}

\section{Related Work}

\subsection{Spatial Intelligence of VLMs}

Spatial intelligence generally refers to a model’s ability to perceive, represent, and reason about object positions, relative relations, directions, distances, and viewpoint changes ~\cite{johnson2017clevr, liu2023visual, yang2025thinking, li2025sti, stogiannidis2503mind}. To study this capability, prior work has examined the spatial performance of VLMs across diverse input modalities, including static images, point clouds, multi-view images, and videos ~\cite{chen2024spatialvlm, xu2024pointllm, li2025viewspatial, li2025llava, lin2025mmsi}. These studies cover a broad range of abilities, such as relative spatial relation understanding, absolute spatial reasoning, object-centric spatial modeling, view correspondence, viewpoint transformation, and temporal spatial memory. Existing analyses consistently show that, although current VLMs have developed strong capabilities in semantic recognition and description, they still struggle with stable spatial representations and explicit spatial reasoning, especially under view variations, long-range temporal dependencies, and complex scene conditions ~\cite{yang2025cambrian, liu2025can, pothiraj2025capture, hong2025motionbench, yuan2025videorefer}. These findings highlight the need for benchmarks that can systematically evaluate the spatial intelligence of VLMs across diverse data sources and scenario settings.

\subsection{Benchmarks for Spatial Intelligence}

Existing benchmarks for spatial intelligence cover tasks such as spatial relation understanding, viewpoint transformation, dynamic scene modeling, and cross-view reasoning ~\cite{yang2025thinking, yeh2025seeing, zhang2025open3d, zha2025aircopbench, zhang2025your, zhao2025urbanvideo, zhou2025urbench}. They evaluate models under diverse scenarios and input modalities, including ground-level scenes, open environments, and multi-source observations. For example, VSI-Bench~\cite{yang2025thinking} and UrbanVideo-Bench~\cite{zhao2025urbanvideo} support dynamic observations but lack cross-view modeling; UrBench ~\cite{zhou2025urbench} and CityCube~\cite{xu2026citycube} include cross-view tasks but are mainly based on static images; and other benchmarks, such as AirCopBench~\cite{zha2025aircopbench}, Open3D-VQA~\cite{zhang2025open3d} and SpatialSky-Bench~\cite{zhang2025your} increase task diversity and scene complexity without explicitly modeling joint UAV-satellite inputs. Collectively, existing benchmarks rarely capture the full combination of capabilities required in dynamic UAV-satellite settings. In contrast, LinkS$^2$Bench takes UAV videos and satellite imagery as joint inputs and systematically evaluates VLMs’ cross-view spatial intelligence under dynamic and cross-view conditions. Table ~\ref{tab:comparison} summarizes the differences between LinkS$^2$Bench and representative existing benchmarks.

\begin{figure}[t]
    \centering
    \subfloat[Distribution of tasks]{\includegraphics[width=0.6\linewidth]{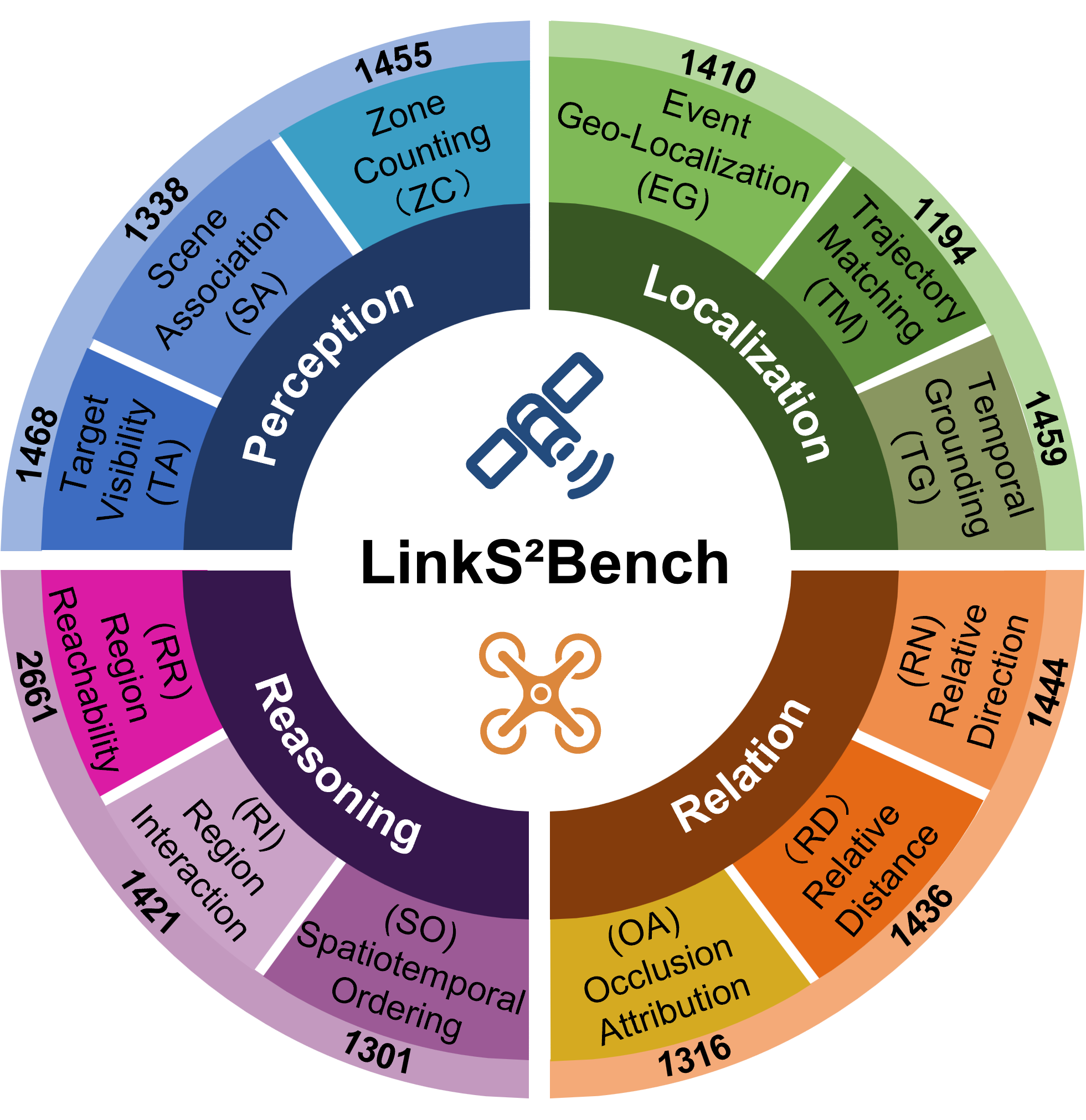}
    \label{fig:paradigm_existing_methods}}
    \hfill
    \subfloat[Distribution of data sources]{\includegraphics[width=0.6\linewidth]{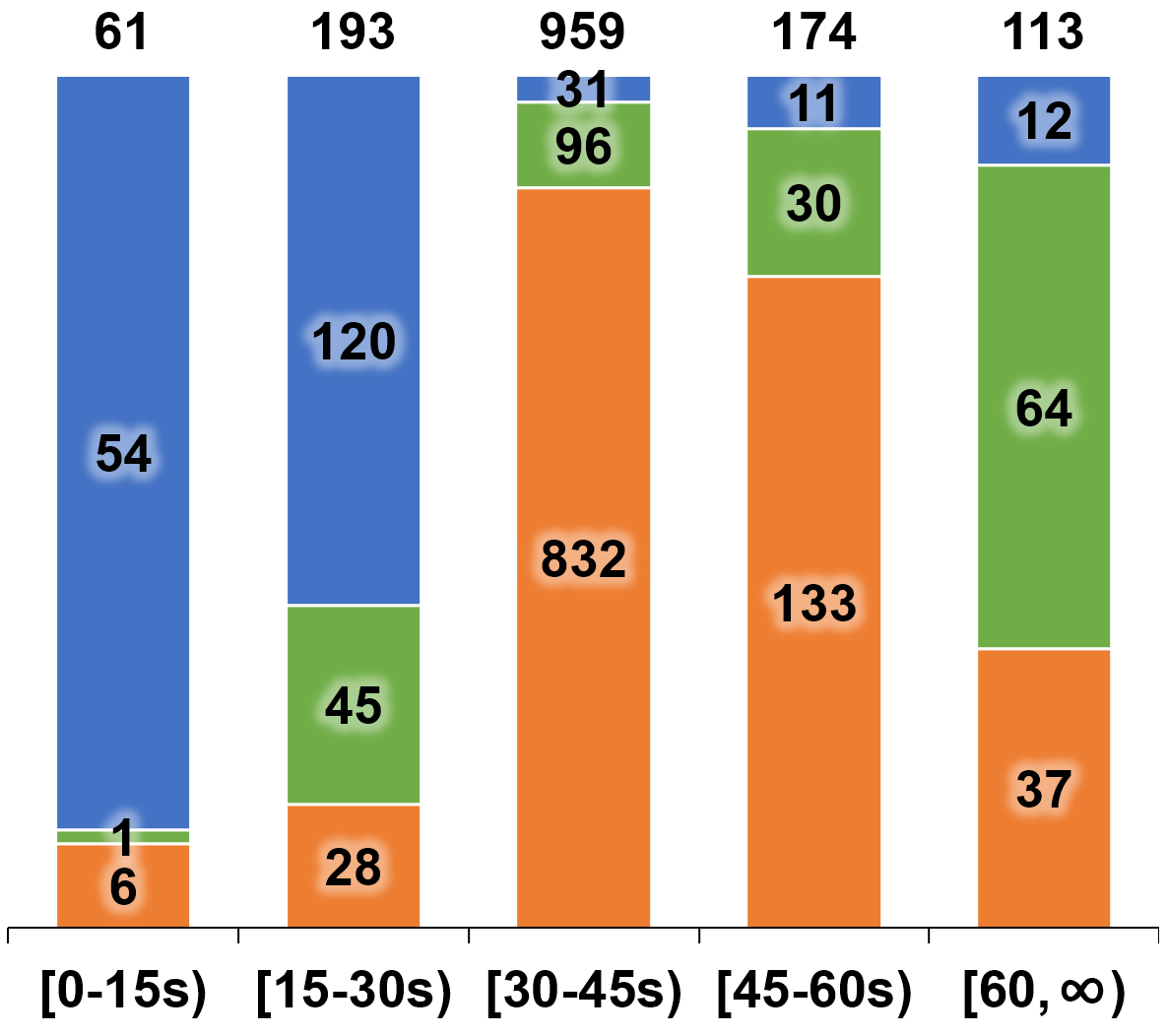}
    \label{fig:paradigm_our_methods}}
    \caption{Benchmark Statistics. (a) The distribution of tasks across four main categories. (b) Distribution of data sources across different video durations.}
    \vspace{-10pt}
    \label{fig:statistics}
\end{figure}

\begin{figure*}[t]
  \centering
  \includegraphics[width=\linewidth]{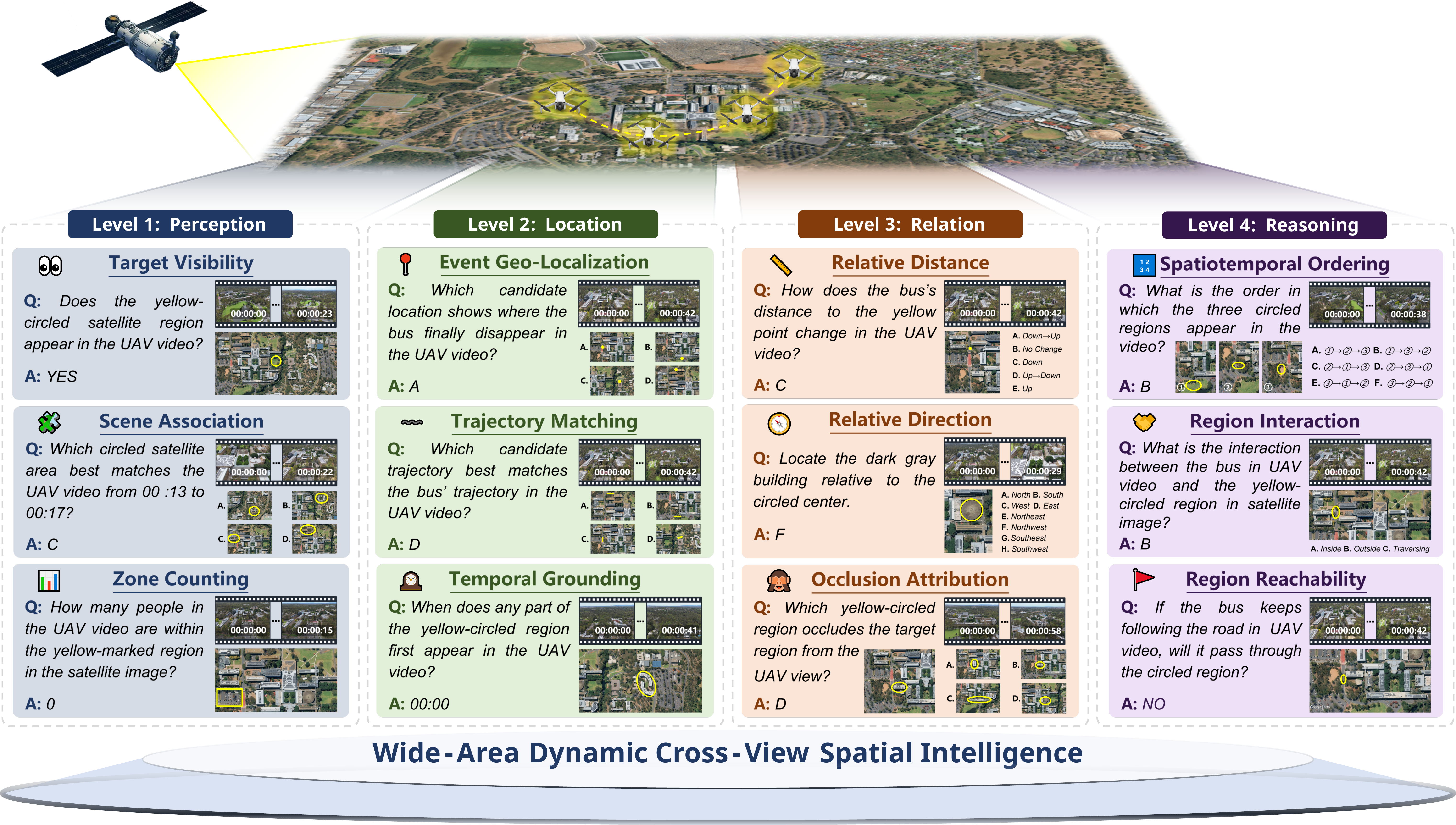}
  \caption{LinkS$^2$Bench  comprises 12 fine-grained task types categorized into four primary evaluation dimensions: Perception, Location, Relation, and Reasoning. For visualization purposes, the satellite images are cropped to highlight the annotations; the actual benchmark samples preserve the full original spatial context.}
  \label{fig:tasks_high_res}
\end{figure*}

\section{LinkS$^2$Bench}

\subsection{Overview}

We introduce LinkS$^2$Bench, a novel benchmark evaluating VLMs' dynamic UAV-satellite cross-view spatial intelligence, critical for applications like autonomous navigation, dynamic monitoring, and wide-area search-and-rescue. As illustrated in Figure ~\ref{fig:statistics}, LinkS$^2$Bench comprises 17,903 QA pairs derived from 1,500 real-world videos (>17 hours) and 43,273 satellite images (>200 km² coverage). The videos span diverse durations (primarily 30–45s) and are sourced via manual collection (69\%), public websites (15\%), and existing datasets (16\%).
LinkS$^2$Bench targets the core challenge of anchoring dynamic local observations to static global maps. To systematically evaluate this, we construct 12 fine-grained tasks across four dimensions: perception, localization, relation, and reasoning.

\vspace{-2pt}

\subsection{Task Design}

LinkS$^2$Bench comprehensively evaluates VLMs in dynamic cross-view scenarios across four dimensions: perception, localization, relation, and reasoning. Figure ~\ref{fig:tasks_high_res} illustrates the specific task types within each dimension.

\textbf{Perception} tasks evaluate whether VLMs can accurately establish cross-view correspondences between drone videos and satellite imagery, forming the basis for dynamic cross-view understanding. Specifically, \textit{Target Visibility} determines whether a marked satellite region appears in the video, \textit{Scene Association} identifies the satellite region corresponding to a specific video segment, and \textit{Zone Counting} estimates the number of vehicles or pedestrians within an annotated satellite region.

\textbf{Localization} tasks examine whether VLMs can temporally and spatially ground dynamic observations from local video views into the global map view. In particular, \textit{Temporal Grounding} predicts the timestamp when a marked satellite region appears in the drone video, \textit{Event Geo-localization} pinpoints the location of target events in the satellite image, and \textit{Trajectory Matching} identifies the annotated satellite trajectory that best aligns with a dynamic target in the video.

\textbf{Relation} tasks focus on whether VLMs can comprehend the spatial relationships between dynamic targets and satellite regions across views. \textit{Relative Distance} assesses variations in the distance between a target and a region, \textit{Relative Direction} evaluates the directional orientation of targets relative to annotated regions, and \textit{Occlusion Attribution} infers whether two marked satellite regions exhibit an occlusion relationship based on video evidence.

\textbf{Reasoning} tasks measure higher-level dynamic cross-view spatial capabilities of VLMs beyond direct perception and grounding. \textit{Region Reachability} determines whether a target can reach a designated region under a predefined future action, \textit{Spatiotemporal Ordering} requires identifying the chronological sequence of multiple satellite regions appearing in the video, and \textit{Region Interaction} evaluates whether a target is actively engaging with an annotated satellite region.

To ensure consistent evaluation across diverse task types, many tasks in LinkS$^2$Bench are formulated as multiple-choice questions. A few tasks adopt specialized answer formats to better match their objectives: \textit{Target Visibility} and \textit{Region Reachability} use binary Yes/No answers, \textit{Zone Counting} requires a numerical response, and \textit{Temporal Grounding} requires predicting a timestamp.

\begin{figure*}[t]
  \centering
  \includegraphics[width=0.9\linewidth]{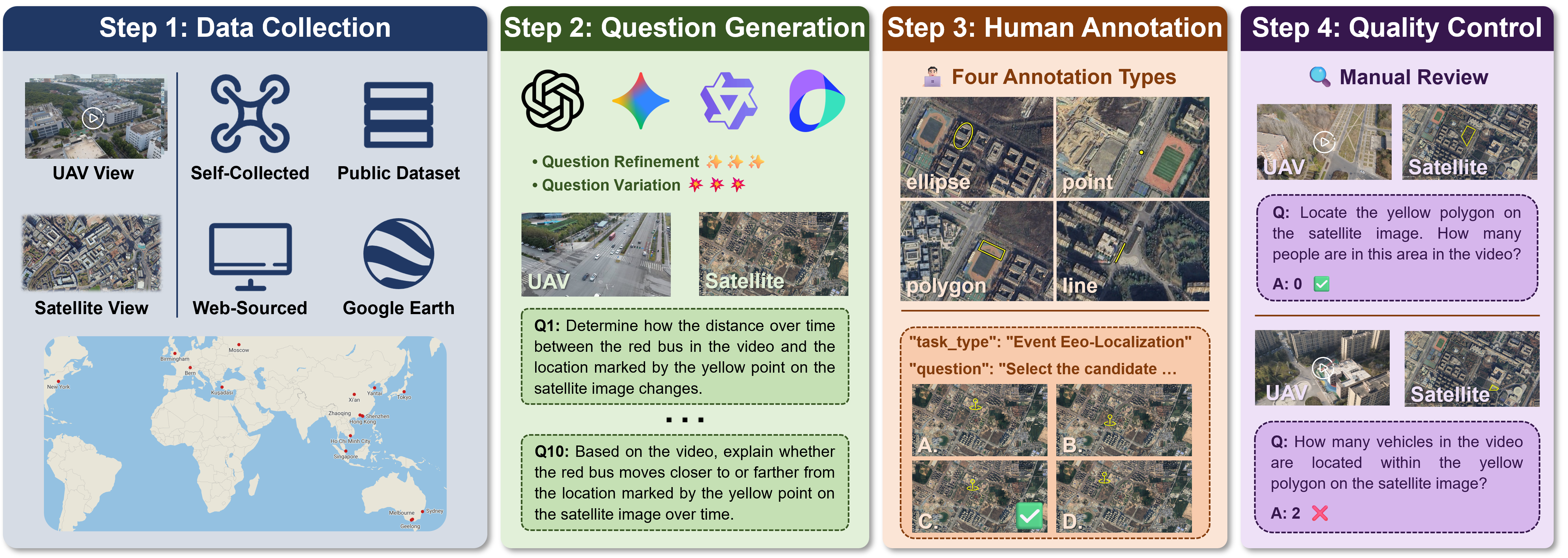}
  \caption{LinkS$^2$Bench curation pipeline includes data collection, question generation, human annotation and quality control.}
  \label{fig:pipeline}
\end{figure*}

\subsection{Benchmark Construction Pipeline}
As illustrated in Figure ~\ref{fig:pipeline}, the benchmark construction pipeline consists of four main stages: data collection, question generation, human annotation, and quality control. Constructing LinkS$^2$Bench involved a large amount of manual effort across all stages, requiring over 1,800 human hours in total, among which human annotation alone took more than 1,000 hours.

\textbf{Data Collection.} We collected UAV flight videos and corresponding satellite imagery from 16 cities worldwide. In particular, the Xi’an data were collected by our team using three DJI Mini 3 drones. To further expand the dataset, we additionally incorporated video data from two cities, Zhaoqing and Shenzhen, from the public UrbanVideoBench ~\cite{zhao2025urbanvideo}. In addition, we supplemented the dataset with UAV videos from diverse regions around the world, sourced from public websites such as Pexels and Pixabay. The satellite images were acquired through Google Earth and then matched to the UAV videos. The overall data collection and construction process, including video acquisition, data cleaning, sample organization, satellite matching, and correspondence verification, required over 800 hours of manual effort.

\textbf{Question Generation.} To ensure the diversity and robustness of the benchmark, we design task-specific question formats for all 12 fine-grained spatial understanding tasks. Specifically, we leverage multiple VLMs to generate 10 different question templates for each task, covering diverse linguistic expressions and query structures, thereby reducing the risk of over-reliance on superficial pattern matching. For the answers, we adopt standardized formats to facilitate automatic evaluation. Specifically, scene association, event geo-localization, trajectory matching, relative direction, relative distance, occlusion attribution, spatiotemporal ordering, and region interaction are formulated as multiple-choice questions with text or image options, while the remaining tasks use open-ended responses, such as binary judgments, short numerical answers, or concise timestamp expressions.

\textbf{Human Annotation.} We employ a semi-automatic framework requiring over 1,000 hours of manual effort (averaging  approximately 4 minutes per sample). Annotators visually ground spatial elements using four formats: ellipses (target instances), points (key locations), polygons (region boundaries), and lines (trajectories). Based on these spatial markings and predefined templates, the system automatically compiles structured QA pairs. To ensure rigorous evaluation, multiple-choice questions are augmented with rule-based distractors. Finally, all generated samples undergo strict manual verification for quality control.

\textbf{Quality Control}. At this stage, we introduce a manual verification procedure to reduce potential biases in the construction of LinkS$^2$Bench. Specifically, multiple annotators review the generated question–answer samples to identify possible annotation errors, such as incorrect multiple-choice options and incorrect open-ended answers. These errors mainly arise from two challenges: the drastic viewpoint changes and limited local fields of view in UAV videos, and the inherent difficulty of cross-view alignment between UAV videos and satellite images. Through this multi-level quality control process, we improve the robustness and annotation accuracy of LinkS$^2$Bench. More details on benchmark construction are provided in the supplementary material.

\section{Experiments}

In this section, we conduct a comprehensive evaluation of 18 representative VLMs on dynamic UAV–satellite cross-view tasks in LinkS$^2$Bench. We then analyze the experimental results from both model and task perspectives to better understand the capabilities and limitations of current VLMs, and further summarize the major causes of model failures across different tasks. Finally, we investigate model adaptation strategies, including cross-view alignment enhancement and supervised fine-tuning, to examine the utility of LinkS$^2$Bench beyond benchmark evaluation.

\subsection{Experimental Setup}

\subsubsection{Benchmark Models} We evaluate a broad range of VLMs covering diverse model families, parameter scales, and training paradigms. For proprietary models, we include four representative model families, namely Gemini-3.1 (Google 2026), GPT-5.4 (OpenAI 2026), Claude-Sonnet-4.6 (Anthropic 2026) and Doubao-seed-1-8 ~\cite{seed2026seed1}. For open-source models, we further examine recent advances represented by Llama-4-Scout ~\cite{adcock2026llama}, Kimi-K2.5 ~\cite{team2026kimi}, GLM-4.6V-Flash ~\cite{hong2025glm}, Ministral-3 ~\cite{liu2026ministral}, Qwen3.5 ~\cite{qwen3.5}, and LLaVA-OneVision-1.5 ~\cite{an2025llava}. In all experiments, we follow a unified standard protocol; unless otherwise specified, the sampling temperature is set to 0.

\subsubsection{Metric Design} Our benchmark contains four categories of questions. For multiple-choice and and yes/no questions, we use Accuracy (ACC) as the primary evaluation metric. For numerical-answer questions, we adopt Mean Relative Accuracy (MRA) as the primary evaluation metric, following the evaluation protocol in ~\cite{yang2025thinking}.  Since our benchmark contains samples with zero-valued ground truth, we extend the original MRA formulation with a piecewise definition to properly handle such cases. Concretely, when the ground-truth value is zero, we apply a dedicated zero-case criterion; otherwise, we compute the relative error according to the standard MRA formulation using thresholds $C=\{0.50, 0.55, \ldots, 0.95\}$:
\begin{equation}
\mathrm{RA}_{\theta}(\hat{y}, y)=
\begin{cases}
\mathbf{1}\!\left(\dfrac{|\hat{y}-y|}{y} < 1-\theta\right), & y>0,\\
\mathbf{1}(\hat{y}=0), & y=0,
\end{cases}
\end{equation}
where $\hat{y}$ and $y$ represent the predicted and ground-truth values, respectively. $\mathbf{1}(\cdot)$ is the indicator function.
\begin{equation}
\mathrm{MRA}(\hat{y}, y)
=
\frac{1}{10}
\sum_{\theta \in C}
\mathrm{RA}_{\theta}(\hat{y}, y),
\end{equation}
For timestamp-answer questions, following SoccerNet ~\cite{deliege2021soccernet}, we report ACC@1s, where a prediction is regarded as correct if its temporal deviation from the ground truth is no greater than 1 second.
\begin{equation}
\mathrm{ACC@1s}(\hat{y}, y)=\mathbf{1}(|\hat{y}-y|\le 1)
\end{equation}

\subsubsection{Human Level Performance.} We construct LinkS$^2$Bench (tiny) by sampling 720 questions from the full benchmark, with 60 questions for each task. Human evaluators answer all questions independently, and their performance is measured using the aforementioned metrics. For comparison, we additionally evaluate Gemini-3.1-Pro, GPT-5.4, Claude-Sonnet-4.6, Doubao-seed-1-8, GLM-4.6V-Flash, and Qwen3.5-397B-A17B on the same subset of LinkS$^2$Bench. More details on the experimental setup on LinkS$^2$Bench-tiny are provided in the supplementary material.

\begin{table*}[t]
\caption{Results (\%) on LinkS$^2$Bench for existing various VLMs on 12 task types across 4 evaluation dimensions. The former section shows existing popular models’ results. The latter section demonstrates model adaption experimental results.}
\label{tab:main_results}
\centering
\footnotesize
\renewcommand{\arraystretch}{1.3}
\setlength{\tabcolsep}{0.5pt}
\resizebox{\textwidth}{!}{%
\begin{tabular}{l @{\hspace{0.2pt}} c @{\hspace{0.5pt}} lllllllllllllllll}
\toprule
\multirow{2}{*}{Method} & \multirow{2}{*}{Rank} & \multirow{2}{*}{Avg.}
& \multicolumn{4}{c}{\cellcolor{myblue}\textbf{Perception}}
& \multicolumn{4}{c}{\cellcolor{mygreen}\textbf{Localization}}
& \multicolumn{4}{c}{\cellcolor{myorange}\textbf{Relation}}
& \multicolumn{4}{c}{\cellcolor{mypurple}\textbf{Reasoning}} \\
& & & TV & SA & ZC & \textbf{Avg.} & EG & TM & TG & \textbf{Avg.} & RN & RD & OA & \textbf{Avg.} & SO & RI & RR & \textbf{Avg.} \\

\midrule

\rowcolor{gray!20}
\multicolumn{19}{l}{\textit{Baseline}} \\
Random      & - &      & 50.0 & 25.0 & - & - & 25.0 & 25.0 & - & - & 12.5 & 20.0 & 25.0 & 19.2  & 16.7 & 33.3 & 50.0 & 33.3 \\

\rowcolor{gray!20}
\multicolumn{19}{l}{\textit{LinkS$^2$Bench (tiny) Perf.}} \\
Human Level     & - & 91.3 & 98.3 & 96.7 & 85.8 & 93.6 &  90.0 & 91.7 & 88.3 & 90.0 & 93.3 & 95.0 & 86.7 & 91.7 & 85.0 & 93.3 & 91.2 & 89.8 \\
Gemini-3.1-Pro  & - & 52.7 & 76.7 & 60.0 & 60.7 & 65.8 & 53.3 & 48.3 & 41.7 & 47.8 & 53.3 & 28.3 & 48.3 & 43.3 & 50.0 & 46.7 & 65.0 & 53.9 \\
GPT-5.4          & - & 48.1 & 85.0 & 61.7 & 69.0 & 71.9 & 46.7 & 45.0 & 45.0 & 45.6 & 23.3 & 33.3 & 38.3 & 31.6 & 18.3 & 56.7 & 55.0 & 43.3 \\
Claude-Sonnet-4.6 & - & 37.8 & 51.7 & 55.0 & 68.2 & 58.3 & 30.0 & 26.7 & 26.7 & 27.8 & 23.3 & 15.0 & 41.7 & 26.7 & 15.0 & 55.0 & 45.0 & 38.3 \\
Doubao-seed-1-8  & - & 47.8 & 76.7 & 70.0 & 40.8 & 62.5 & 55.0 & 41.7 & 43.3 & 46.7 & 25.0 & 28.3 & 38.3 & 30.5 & 41.7 & 48.3 & 65.0 & 51.7 \\
GLM-4.6V-Flash & - & 45.0 & 85.0 & 56.7 & 70.0 & 70.6 & 41.7 & 38.3 & 15.0 & 31.7 & 16.7 & 26.7 & 30.0 & 24.5 & 40.0 & 51.7 & 68.3 & 53.3 \\
Qwen3.5-397B-A17B & - & 46.7 & 76.7 & 66.7 & 63.3 & 68.9 & 40.0 & 46.7 & 20.0 & 35.6 & 20.0 & 23.3 & 43.3 & 28.9 & 36.7 & 53.3 & 70.0 & 53.3 \\

\rowcolor{gray!20}
\multicolumn{19}{l}{\textit{Proprietary Models (API)}} \\
Gemini-3.1-Flash & 4 & 42.8 & \textbf{84.7} & 47.0 & \textbf{68.8} & 66.8 & 35.3 & 40.7 & 49.3 & 41.8 & 15.7 & 25.0 & 25.7 & 22.1 & 11.3 & 54.7 & 55.7 & 40.6 \\
Gemini-3.1-Pro   & \multicolumn{1}{c}{\cellcolor{1}\textbf{1}} & \textbf{51.1} & 84.4 & 57.3 & 63.7 & 68.5 & 44.1 & 44.7 & \textbf{51.9} & \textbf{46.9} & \textbf{29.9} & 30.6 & \textbf{47.3} & \textbf{35.9} & \textbf{44.0} & \textbf{57.3} & 58.1 & \textbf{53.1} \\
GPT-5.4-mini     & 5 & 38.2 & 76.3 & 39.3 & 54.4 & 56.7 & 30.0 & 33.0 & 28.7 & 30.6 & 14.0 & \textbf{33.3} & 25.3 & 24.2 & 21.7 & 44.7  & 58.0 & 41.5 \\
GPT-5.4          & \multicolumn{1}{c}{\cellcolor{3}\textbf{3}} & 46.6 & 81.4 & \textbf{66.8} & 63.6 & \textbf{70.6} & 44.0 & 43.0 & 44.6 & 43.9 & 24.4 & 29.8 & 37.1 & 30.4 & 16.4 & 50.1 & 57.6 & 41.4 \\
Claude-Sonnet-4.6   & 6 & 37.7 & 53.3 & 43.3 & 67.3 & 54.6 & 32.7 & 31.3 & 27.3 & 30.4 & 21.3 & 19.0 & 41.0 & 27.1 & 13.7 & 54.0 & 48.0  & 38.6 \\
Doubao-seed-1-8  & \multicolumn{1}{c}{\cellcolor{2}\textbf{2}} & 47.8 & 76.7 & 66.7 & 41.1 & 61.5 & \textbf{50.3} & \textbf{47.3} & 42.3 & 46.6 & 23.3 & 29.0 & 40.7 & 31.0 & 42.7 & 50.3 & \textbf{63.7} & 52.2 \\

\rowcolor{gray!20}
\multicolumn{19}{l}{\textit{Open-source Models}} \\
Llama-4-Scout          & 8 & 26.6 & 79.6 & 10.4 & 22.5 & 37.5 & 4.1 & 22.6 & 20.4 & 15.7 & 12.2 & 16.3 & 14.3 & 14.3 & 31.3 & 24.5 & 60.4  & 38.7 \\
Kimi-K2.5              & 4 & 38.4 & 55.1 & 51.0 & 73.5 & 59.9 & 32.7 & 26.1 & \textbf{48.0} & 35.6 & 20.4 & 17.4 & 27.3 & 21.7 & 25.0 & 43.8 & 40.0 & 36.3 \\
GLM-4.6V-Flash           & \multicolumn{1}{c}{\cellcolor{2}\textbf{2}} & 44.3 & \textbf{85.9} & 57.1 & 71.1 & \textbf{71.4} & \textbf{39.7} & 38.7 & 14.3 & 30.9 & 14.8 & 23.5 & 32.5 & 23.6 & 42.1 & 50.0 & 62.2 & 51.4 \\
Ministral-3-3B         & 11 & 22.9 & 17.7 & 11.9 & 61.9 & 30.5 & 16.9 & 14.5 & 20.2 & 17.2 & 13.8 & 18.5 & 18.9 & 17.1 & 14.9 & 27.5 & 37.8 & 26.7 \\
Ministral-3-8B         & 11 & 22.9 & 15.3 & 8.5 & 59.1 & 27.6 & 13.0 & 19.3 & 27.4 & 19.9 & 16.1 & 14.8 & 12.8 & 14.6 & 12.5 & 37.5 & 38.9 & 29.6 \\
Ministral-3-14B        & 10 & 23.3 & 13.9 & 7.7 & 64.1 & 28.6 & 15.0 & 18.2 & 26.3 & 19.8 & 13.0 & 13.5 & 17.9 & 14.8 & 12.5 & 41.4 & 35.8 & 29.9 \\
Qwen3.5-4B             & 9 & 23.7 & 27.0 & 5.3 & 63.7 & 32.0 & 8.7 & 7.3 & 32.1 & 16.0 & 12.1 & 23.8 & 26.4 & 20.8 & 6.5 & 35.6 & 36.0 & 26.0 \\
Qwen3.5-9B             & 7 & 30.7 & 26.3 & 19.0 & \textbf{74.8} & 40.0 & 18.0 & 15.3 & 36.8 & 23.4 & 16.0 & 22.7 & 36.0 & 24.9 & 17.3 & 48.5 & 37.3 & 34.4 \\
Qwen3.5-35B-A3B            & \multicolumn{1}{c}{\cellcolor{3}\textbf{3}} & 44.1 & 71.4 & 54.2 & 42.6 & 56.1 & 29.8 & 42.6 & 36.7 & 36.4 & \textbf{38.3} & 31.9 & \textbf{41.7} & \textbf{37.3} & 34.7 & 46.8 & 58.3  & 46.6 \\
Qwen3.5-397B-A17B      & \multicolumn{1}{c}{\cellcolor{1}\textbf{1}} & \textbf{45.6} & 66.7 & \textbf{73.3} & 36.1 & 58.7 & 31.0 & \textbf{50.0} & 39.5 & \textbf{40.2} & 23.9 & \textbf{37.5} & 41.2 & 34.2 & 26.1 & \textbf{55.1} & \textbf{66.7} & 49.3 \\
LLaVA-OneVision-1.5-4B & 6 & 33.2 & 64.7 & 16.7 & 53.3 & 44.9 & 20.2 & 20.0 & 17.9 & 19.4 & 13.4 & 18.5 & 10.4 & 14.1 & \textbf{52.6} & 47.5 & 63.7 & \textbf{54.6} \\
LLaVA-OneVision-1.5-8B & 5 & 33.7 & 54.1 & 24.4 & 61.5 & 46.7 & 22.8 & 24.0 & 34.5 & 27.1 & 12.2 & 27.2 & 18.1 & 19.2 & 23.7 & 40.0 & 61.5 & 41.7 \\

\rowcolor{gray!20}
\multicolumn{19}{l}{\textit{Fine-tuning Experiments}} \\
Qwen3.5-4B & - & 28.2 & 21.0 & 5.1 & 71.6 & 32.6 & 9.1 & 15.1 & 70.3 & 31.5 & 11.2 & 29.3 & 21.0 & 20.5 & 14.7 & 30.7 & 38.8 & 28.1 \\
Qwen3.5-4B$_{\text{\tiny-Finetuned}}$ & - & 52.6$_{\text{\tiny+24.4}}$ & 87.2$_{\text{\tiny+66.2}}$ & 75.9$_{\text{\tiny+70.8}}$ & 76.4$_{\text{\tiny+4.8}}$ & 79.8$_{\text{\tiny+47.2}}$ & 50.8$_{\text{\tiny+41.7}}$ & 46.3$_{\text{\tiny+31.2}}$ & 65.4 & 54.2$_{\text{\tiny+22.7}}$ & 12.5$_{\text{\tiny+1.3}}$ & 24.0 & 29.8$_{\text{\tiny+8.8}}$ & 22.1$_{\text{\tiny+1.6}}$ & 62.5$_{\text{\tiny+47.8}}$ & 43.9$_{\text{\tiny+13.2}}$ & 56.5$_{\text{\tiny+17.7}}$ & 54.3$_{\text{\tiny+26.2}}$ \\
Qwen3.5-9B & - & 34.7 & 26.0 & 21.9 & 75.8 & 41.2 & 20.6 & 25.7 & 66.5 & 37.6 & 14.2 & 23.7 & 33.3 & 23.7 & 17.7 & 46.1 & 45.1 & 36.3 \\
Qwen3.5-9B$_{\text{\tiny-Finetuned}}$ & - & 53.1$_{\text{\tiny+18.4}}$ & 87.7$_{\text{\tiny+61.7}}$ & 75.9$_{\text{\tiny+54.0}}$ & 76.7$_{\text{\tiny+0.9}}$ & 80.1$_{\text{\tiny+38.9}}$ & 50.8$_{\text{\tiny+30.2}}$ & 46.3$_{\text{\tiny+20.6}}$ & 69.6$_{\text{\tiny+3.1}}$ & 55.6$_{\text{\tiny+18.0}}$ & 11.1 & 29.2$_{\text{\tiny+5.5}}$ & 29.5 & 23.3 & 58.0$_{\text{\tiny+40.3}}$ & 42.0 & 60.7$_{\text{\tiny+15.6}}$ & 53.6$_{\text{\tiny+17.3}}$ \\

\rowcolor{gray!20}
\multicolumn{19}{l}{\textit{Alignment Adapter Experiments}} \\
Gemini-3.1-Pro w/ CVAA & - & 55.4$_{\text{\tiny+4.3}}$ & 94.9$_{\text{\tiny+10.5}}$ & 58.8$_{\text{\tiny+1.5}}$ & 67.7$_{\text{\tiny+4.0}}$ & 73.8$_{\text{\tiny+5.3}}$ & 50.0$_{\text{\tiny+5.9}}$ & 47.4$_{\text{\tiny+2.7}}$ & 53.9$_{\text{\tiny+2.0}}$ & 50.4$_{\text{\tiny+3.5}}$ & 36.4$_{\text{\tiny+6.5}}$ & 36.6$_{\text{\tiny+6.0}}$ & 43.2 & 38.7$_{\text{\tiny+2.8}}$ & 48.3$_{\text{\tiny+4.3}}$ & 60.6$_{\text{\tiny+3.3}}$ & 66.7$_{\text{\tiny+8.6}}$ & 58.5$_{\text{\tiny+5.4}}$ \\
GPT-5.4 w/ CVAA & - & 51.0$_{\text{\tiny+4.4}}$ & 78.4 & 66.7 & 75.9$_{\text{\tiny+12.3}}$ & 73.7$_{\text{\tiny+3.1}}$ & 36.1 & 56.3$_{\text{\tiny+13.3}}$ & 59.5$_{\text{\tiny+14.9}}$ & 50.6$_{\text{\tiny+6.7}}$ & 24.1 & 35.3$_{\text{\tiny+5.5}}$ & 48.5$_{\text{\tiny+11.4}}$ & 36.0$_{\text{\tiny+5.6}}$ & 21.2$_{\text{\tiny+4.8}}$ & 43.8 & 66.7$_{\text{\tiny+9.1}}$ & 43.9$_{\text{\tiny+2.5}}$ \\
Qwen3.5-4B$_{\text{\tiny-Finetuned}}$ w/ CVAA & - & 54.3$_{\text{\tiny+1.7}}$ & 88.6$_{\text{\tiny+1.4}}$ & 82.5$_{\text{\tiny+6.6}}$ & 88.4$_{\text{\tiny+12.0}}$ & 86.5$_{\text{\tiny+6.7}}$ & 55.6$_{\text{\tiny+4.8}}$ & 45.2 & 67.5$_{\text{\tiny+2.1}}$ & 56.1$_{\text{\tiny+1.9}}$ & 9.3 & 24.4$_{\text{\tiny+0.4}}$ & 30.2$_{\text{\tiny+0.4}}$ & 21.3 & 68.2$_{\text{\tiny+5.7}}$ & 41.9 & 50.0 & 53.4 \\
Qwen3.5-9B$_{\text{\tiny-Finetuned}}$ w/ CVAA & - & 55.0$_{\text{\tiny+1.9}}$ & 87.0 & 80.0$_{\text{\tiny+4.1}}$ & 85.1$_{\text{\tiny+8.4}}$ & 84.0$_{\text{\tiny+3.9}}$ & 53.2$_{\text{\tiny+2.4}}$ & 46.7$_{\text{\tiny+0.4}}$ & 71.7$_{\text{\tiny+2.1}}$ & 57.2$_{\text{\tiny+1.6}}$ & 10.4 & 25.5 & 31.9$_{\text{\tiny+2.4}}$ & 22.6 & 61.2$_{\text{\tiny+3.2}}$ & 41.3 & 66.0$_{\text{\tiny+5.3}}$ & 56.2$_{\text{\tiny+2.6}}$ \\

\bottomrule
\end{tabular}
\label{tab:results}
}
\end{table*}

\subsection{Model Comparison}
We report the performance of VLMs on LinkS$^2$Bench, including task-wise accuracies as well as the average accuracy for each dimension. Based on the quantitative results presented in Table~\ref{tab:results}, we draw the following findings:

\textbf{LinkS$^2$Bench Remains Challenging for Current VLMs.} Even the best proprietary model, Gemini-3.1-Pro, attains an average accuracy of only 51.1\%, while the strongest open-source model, Qwen3.5-397B-A17B, reaches only 45.6\%. More importantly, most evaluated models perform substantially worse, with many open-source models achieving average accuracies below 40. This trend is consistent across different model families and parameter scales, suggesting that the challenge of LinkS$^2$Bench is broad rather than limited to a particular type of VLM. Moreover, current VLMs still fall substantially short of human-level performance (\textbf{91.3\%}), with even the strongest proprietary model lagging behind by nearly \textbf{40} points. Together, these results underscore the value of LinkS$^2$Bench and suggest that dynamic cross-view understanding in UAV–satellite scenarios remains far from being solved.

\textbf{Current VLMs Exhibit Highly Imbalanced Spatial Capabilities.} Rather than failing uniformly across all task groups, current VLMs show highly unbalanced performance on LinkS$^2$Bench. For example, Gemini-3.1-Pro achieves \textbf{68.5\%} on Perception and \textbf{53.1\%} on Reasoning, but only \textbf{46.9\%} on Localization and \textbf{35.9\%} on Relation. These results suggest that dynamic cross-view spatial intelligence is not a single unified ability, but a composition of heterogeneous capabilities, some of which remain substantially less mature than others in current VLMs.

\textbf{Fine-Grained Cross-View Relation Modeling Is More Difficult Than Higher-Level Reasoning}. Compared with Reasoning, Relation task is consistently the weakest aspect for top-performing models. Gemini-3.1-Pro obtains \textbf{53.1\%} on Reasoning but only \textbf{35.9\%} on Relation, while Qwen3.5-397B-A17B achieves \textbf{49.3\%} on Reasoning versus \textbf{34.2\%} on Relation. A similar pattern can be observed at the task level: both models perform relatively better on \textit{Region Interaction} and \textit{Region Reachability}, but remain much weaker on \textit{Relative Distance} and \textit{Relative Direction}. This suggests that current VLMs can sometimes infer coarse spatiotemporal outcomes from high-level cues, yet still struggle to model precise spatial relations between dynamic targets and satellite regions across views.

\begin{figure}[t]
  \centering
  \includegraphics[width=0.8\linewidth]{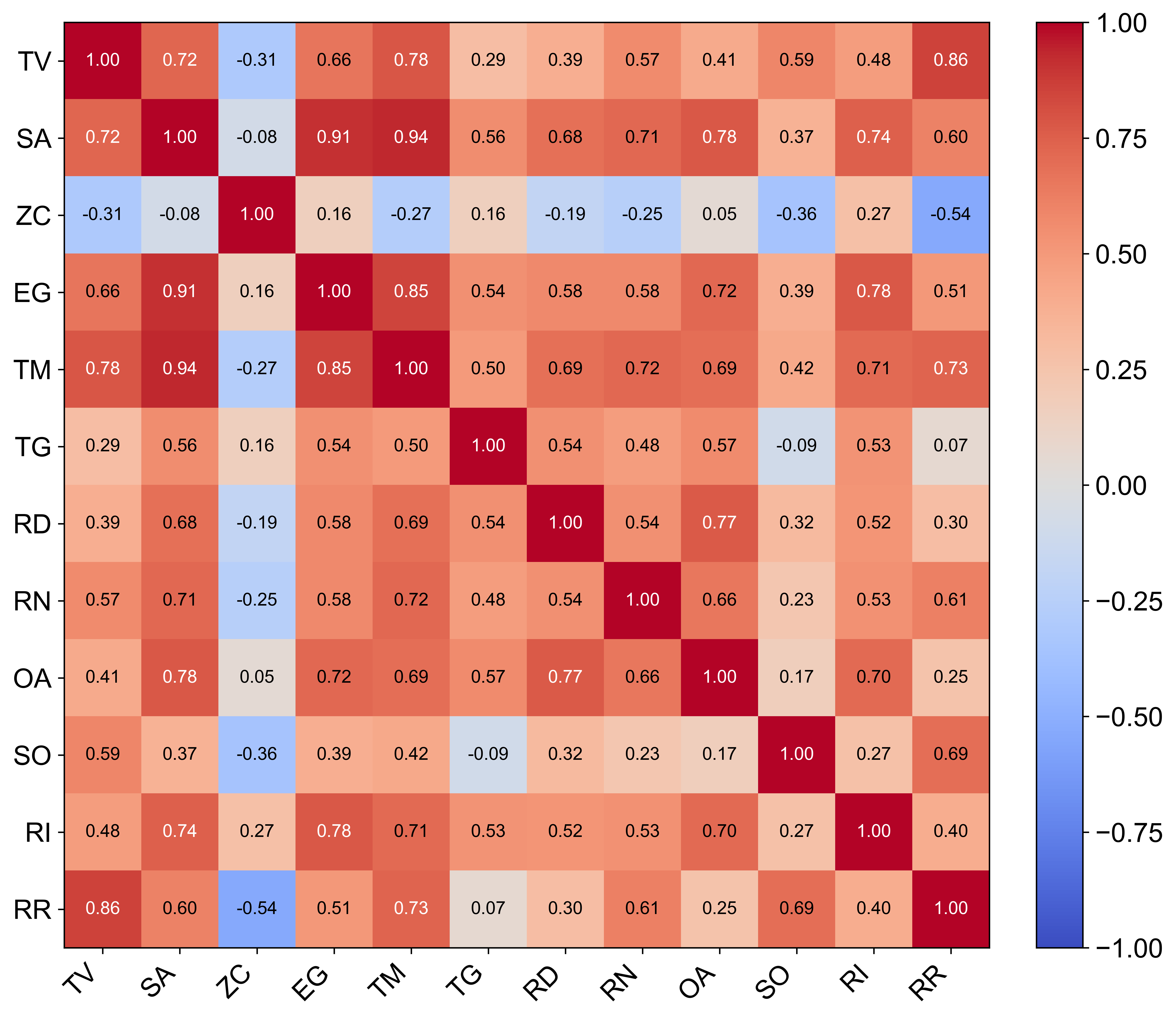}
  \caption{Pairwise task correlations based on VLM performance. Higher values suggest that VLMs perform more similarly on the two tasks, indicating a greater overlap in the abilities required to solve them.}
  \label{fig:correlation}
  \vspace{-9pt}
\end{figure}

\subsection{Correlation Analysis}
To explore task relationships and their underlying capability structures, we compute pairwise correlations of model accuracies based on Table~\ref{tab:results} (excluding LinkS$^2$Bench-tiny and model adaptation results). We assume that highly correlated performance between two tasks indicates shared underlying capabilities. From the resulting correlation matrix in Figure ~\ref{fig:correlation}, we observe the following:

\textbf{LinkS$^2$Bench Evaluates a Diverse Range of Shared and Specialized Capabilities.}
The correlation matrix reveals distinct task clusters alongside significant outliers, highlighting the benchmark's structural diversity. Specifically, tasks like \textit{Scene Association}, \textit{Event Geo-Localization}, and \textit{Trajectory Matching} show strong mutual correlations (up to 0.94), indicating a reliance on shared underlying abilities. Conversely, \textit{Zone Counting} emerges as a highly independent outlier, exhibiting weak or negative correlations with other tasks (e.g., -0.31 with \textit{Target Visibility}). This divergence demonstrates solving LinkS$^2$Bench is inherently multi-faceted, requiring a blend of both shared competencies and highly specialized skills.

\textbf{Localization Serves as the Foundation for Almost All Other Tasks.}
Localization-related tasks exhibit strong correlations across various task categories, underscoring their central role. In dynamic cross-view scenarios, accurate spatial alignment and localization form the critical foundation for subsequent perception, relational understanding, and complex reasoning. 

\subsection{Error Analysis}
To better understand the main bottlenecks of the best-performing VLMs on LinkS$^2$Bench, we conduct a fine-grained error analysis on LinkS$^2$Bench (tiny). Specifically, we group model errors into three distinct categories, informed by both the core visual-spatial capabilities covered by the benchmark and the recurring failure patterns observed in model outputs:

\begin{figure}[t]
  \centering
  \includegraphics[width=0.85\linewidth]{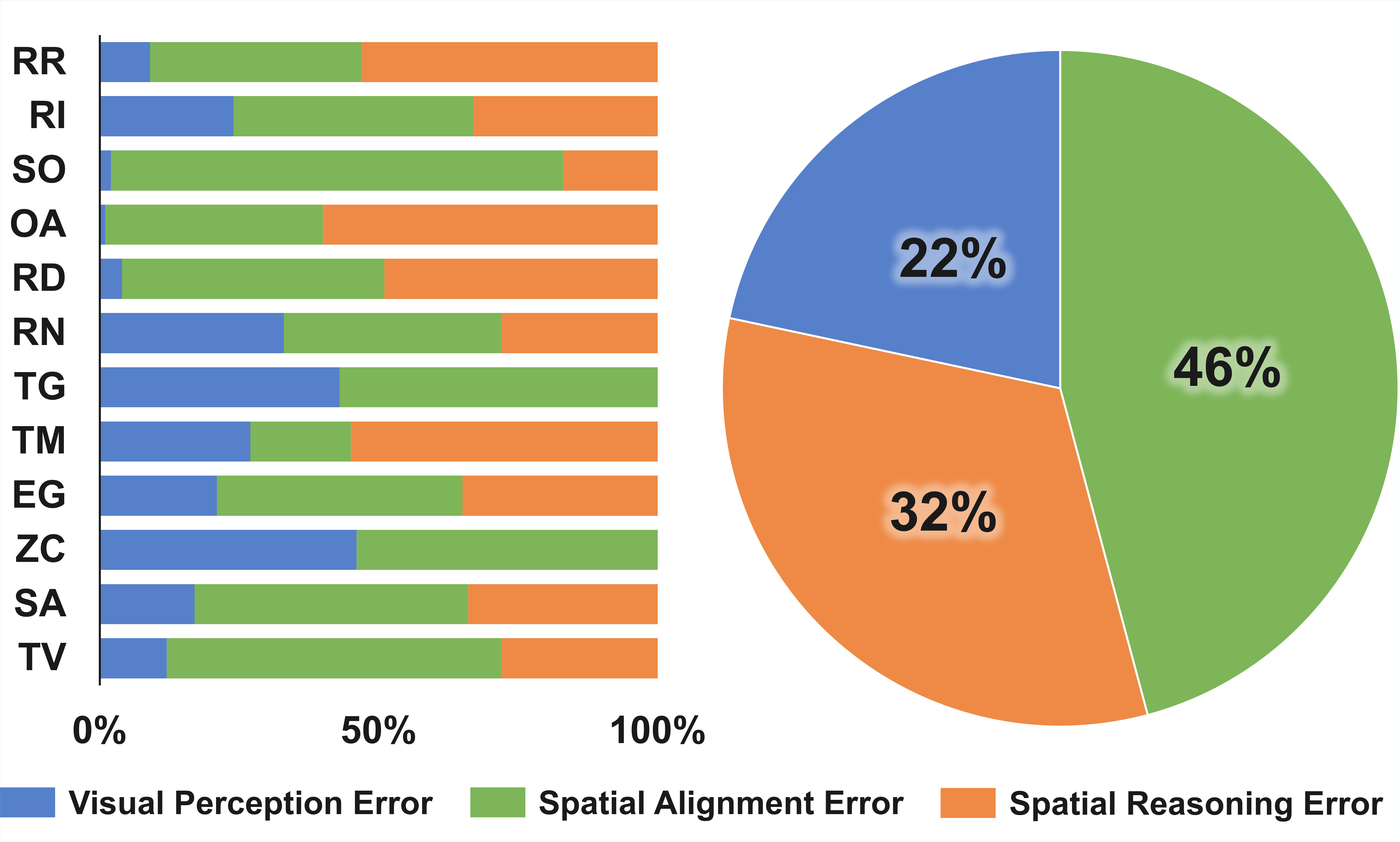}
  \caption{Distribution of error types via manual inspection.}
  \label{fig:error}
\end{figure}

\textbf{Visual Perception Error.} VLMs fail to accurately recognize the objects or dynamic targets described in the question, making it difficult to establish reliable visual correspondences and consequently affecting subsequent spatial localization, relation understanding, and reasoning. \textbf{Spatial Alignment Error.} VLMs struggle with dynamic cross-view spatial alignment, making it difficult to accurately map dynamic observations from drone videos onto the global reference frame of satellite imagery, thereby leading to errors in cross-view correspondence, spatial localization, and dynamic target association.
\textbf{Spatial Reasoning Error.} VLMs face challenges in dynamic cross-view spatial reasoning, making it difficult to jointly model temporal changes and spatial relationships, thereby resulting in errors in higher-level spatiotemporal understanding and reasoning. More detailed failure cases and analyses are provided in the supplementary material.

As shown in Figure ~\ref{fig:error}, spatial alignment errors account for the largest proportion of failures, comprising around \textbf{46\%} of all errors, followed by spatial reasoning errors (32\%) and visual perception errors (22\%). This result suggests that \textbf{dynamic cross-view spatial alignment is the primary bottleneck} for current VLMs on LinkS$^2$Bench. Moreover, alignment errors are broadly observed across multiple tasks, indicating that establishing reliable correspondences between UAV observations and satellite regions remains a fundamental challenge. Building on this observation, we further explore targeted alignment enhancement and supervised fine-tuning to investigate the utility of LinkS$^2$Bench for diagnosing and improving VLMs.

\subsection{Model Adaptation}
In this section, we investigate the utility of LinkS$^2$Bench beyond benchmark evaluation. Specifically, we study whether LinkS$^2$Bench can support both targeted model adaptation and supervised fine-tuning for improving VLMs. To this end, we first introduce a cross-view alignment adapter to alleviate the dominant alignment bottleneck revealed by the error analysis, and then examine supervised fine-tuning on LinkS$^2$Bench to evaluate its value for model adaption.

\begin{figure}[t]
  \centering
  \includegraphics[width=\linewidth]{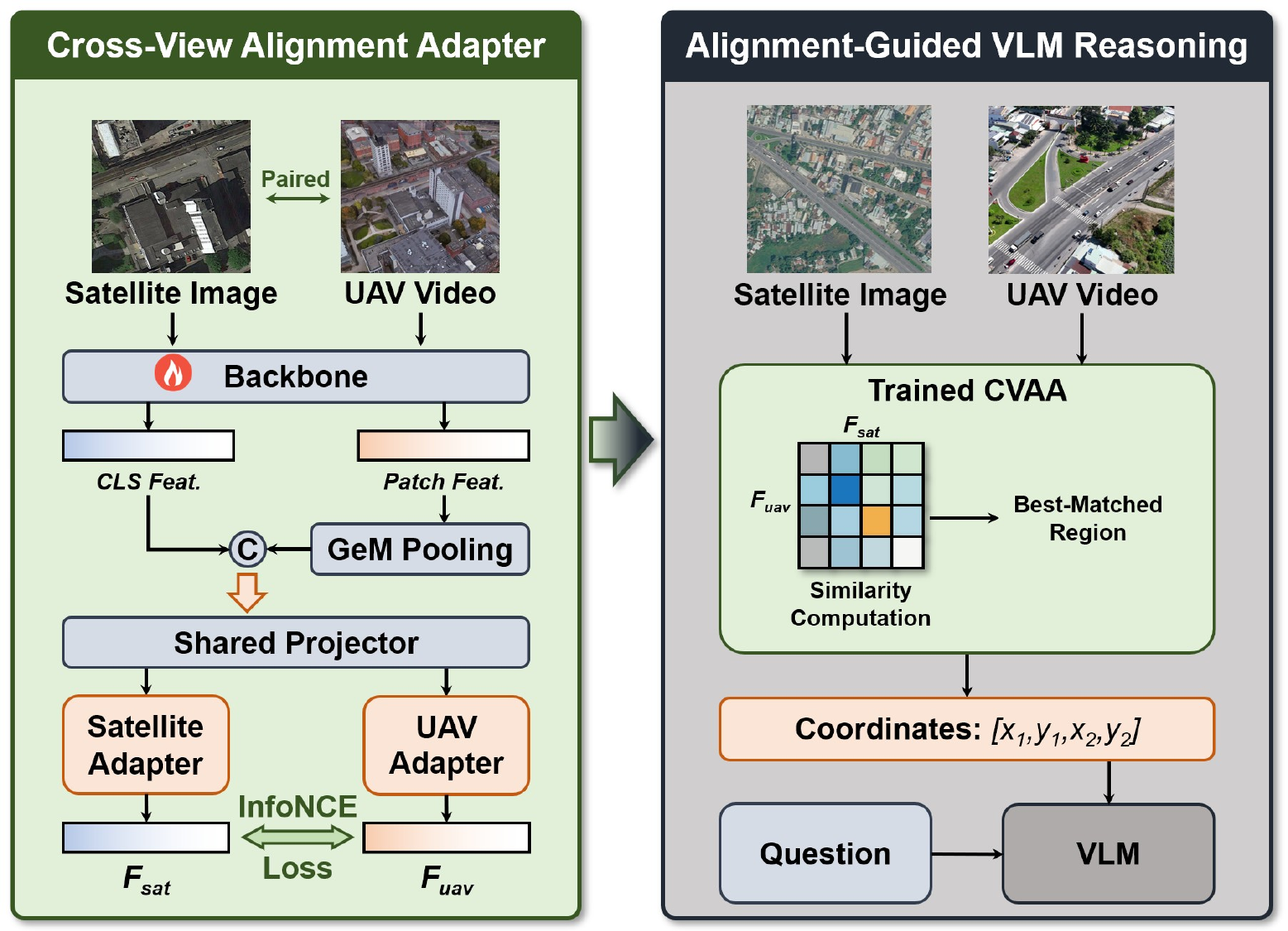}
  \caption{Overview of Cross-View Alignment Adapter and Alignment-Guided VLM Reasoning.
  \vspace{-10pt}
}
  \label{fig:cvaa}
\end{figure}

\subsubsection{Cross-view Alignment Adapter}

Motivated by the predominance of spatial misalignment errors, we introduce the CVAA to provide explicit alignment cues between UAV videos and satellite imagery. Such cross-view alignment is essential because UAV--satellite spatial reasoning cannot be reliably solved from either view alone, but instead requires grounding dynamic local UAV observations in static global satellite context. As illustrated in Figure~\ref{fig:cvaa} (left), CVAA learns cross-view correspondences through a dual-branch architecture under a contrastive learning paradigm.

Specifically, given a satellite image $I_{sat}$ and a UAV image $I_{uav}$, we employ a shared backbone to extract features. For the satellite image, we utilize the CLS token. For the UAV frame, patch-level features are aggregated via Generalized Mean (GeM) pooling. These distinct features are concatenated to form the joint input representation $\mathbf{h}$:
\begin{equation}
    \mathbf{h} = \left[ \mathcal{B}_{cls}(I_{sat}), \text{GeM}(\mathcal{B}_{patch}(I_{uav})) \right]
\end{equation}
Here, $\mathcal{B}_{cls}(\cdot)$ denotes the extraction of the global class token from the shared backbone, $\mathcal{B}_{patch}(\cdot)$ represents the dense patch-level features extracted from the UAV frame. $\mathbf{h}$ is then fed into a shared projector $\mathcal{P}$. Subsequently, modality-specific adapters refine the projected embeddings. The final representations are formulated as:
\begin{equation}
    \mathbf{f}_{sat} = \mathcal{A}_{sat}(\mathcal{P}(\mathbf{h})), \quad
    \mathbf{f}_{uav} = \mathcal{A}_{uav}(\mathcal{P}(\mathbf{h})),
\end{equation}
where $\mathcal{A}_{sat}$ and $\mathcal{A}_{uav}$ represent the satellite and UAV adapters, respectively, and $\mathbf{f}_{sat}$ and $\mathbf{f}_{uav}$ correspond to the final aligned embeddings used for contrastive learning. We employ the InfoNCE loss to maximize the similarity of matched pairs within the batch.

During inference (Figure ~\ref{fig:cvaa}, right), the trained CVAA functions as a retriever. The satellite image is partitioned into overlapping sliding-window candidates. CVAA computes the similarity between the UAV frame and each candidate based on their refined projected features, $\mathbf{f}_{uav}$ and $\mathbf{f}_{sat}$, to identify the best-matched region. The resulting bounding box coordinates $[x_1, y_1, x_2, y_2]$ serve as explicit spatial cues. These coordinates are serialized and concatenated with the user question, providing the VLM with a global alignment prior to ground local UAV observations within the satellite context.

As shown in Table~\ref{tab:results}, CVAA consistently improves the overall performance of strong proprietary VLMs, raising Gemini-3.1-Pro \textbf{from 51.1\% to 55.4\%} and GPT-5.4 \textbf{from 46.6\% to 51.0\%}. More importantly, the gains are most pronounced on localization-related tasks. In particular, the localization average improves to 50.4\% (\textbf{+3.5\%}) for Gemini-3.1-Pro and  to 50.6\% (\textbf{+6.7\%}) for GPT-5.4, with especially notable gains on trajectory matching and temporal grounding. We also observe partial improvements on relation and reasoning tasks, suggesting that stronger cross-view alignment can provide a more reliable spatial basis for downstream dynamic spatial understanding. 

\subsubsection{Supervised Fine-Tuning}
We further study whether our benchmark can serve as effective supervision for adapting VLMs. We conduct supervised fine-tuning on Qwen3.5-4B and Qwen3.5-9B. As shown in Table~\ref{tab:main_results}, both models benefit substantially from fine-tuning, with the overall average reaching 52.6\% (\textbf{+24.4\%}) for Qwen3.5-4B and 53.1\% (\textbf{+18.4\%}) for Qwen3.5-9B. The gains are particularly evident in perception.  At the task level, especially large improvements are observed on target visibility, scene association, and event geo-localization. These results suggest that LinkS$^2$Bench is valuable not only as a diagnostic benchmark, but also as an effective source of supervision for improving VLMs. Moreover, combining the fine-tuned models with CVAA yields further gains, increasing the overall performance \textbf{from 52.6\% to 54.3\%} for Qwen3.5-4B-Finetuned and \textbf{from 53.1\% to 55.0\%} for Qwen3.5-9B-Finetuned. This suggests that explicit cross-view alignment cues remain complementary even after supervised adaptation. Additional results and settings are provided in the supplementary material.

\section{Conclusion}
We present LinkS$^2$Bench, a benchmark for dynamic UAV-satellite cross-view spatial intelligence, built on 17.9k high-quality VQA pairs from real UAV videos and paired satellite imagery. LinkS$^2$Bench enables systematic evaluation of VLMs in large-scale spatial localization, dynamic cross-view perception, and spatiotemporal reasoning. Experiments on 18 representative VLMs show that current models remain far behind human performance, and reveal cross-view dynamic alignment as a major bottleneck. To address this issue, we introduce CVAA, which improves performance by providing explicit alignment cues, and further show through supervised fine-tuning that LinkS$^2$Bench is also valuable as a source of supervision for model adaptation. We expect LinkS$^2$Bench to provide a useful foundation for future research on dynamic cross-view reasoning and multimodal spatial intelligence.                     

\bibliographystyle{ACM-Reference-Format}
\bibliography{samples/sample-base}

\end{document}